\documentclass[runningheads]{llncs}
\usepackage{graphicx}
\usepackage{amssymb,amsfonts}%
\usepackage{mathrsfs}%
\usepackage{xcolor}%
\usepackage{textcomp}%
\usepackage{manyfoot}%
\usepackage{booktabs}%
\usepackage{algorithm}%
\usepackage{algorithmicx}%
\usepackage{algpseudocode}%
\usepackage{listings}%
\usepackage{color}%
\usepackage{float}%
\usepackage{hyperref}%

\usepackage{url}%
\usepackage{optidef}%
\usepackage{multirow}%
\usepackage{lscape}
\usepackage{subcaption}
\usepackage{comment}
\usepackage{enumitem}

\begin{document}

\title{The Influence of Neural Networks on Hydropower Plant Management in Agriculture: Addressing Challenges and Exploring Untapped Opportunities}

\titlerunning{The Influence of NNs on Hydropower Plant Management in Agriculture}

\author{C. Coelho\inst{1}\orcidID{0009-0009-4502-937X} \and
M. Fernanda P. Costa\inst{1}\orcidID{0000-0001-6235-286X} \and
L.L. Ferrás\inst{1,2}\orcidID{0000-0001-5477-3226}}
\authorrunning{C. Coelho et al.}
\institute{Centre of Mathematics (CMAT), University of Minho, Braga, Portugal
\email{cmartins@cmat.uminho.pt, mfc@math.uminho.pt}\\
\and
Department of Mechanical Engineering (Section of Mathematics) - FEUP, University of Porto, Porto, Portugal\\
\email{lferras@fe.up.pt}}
\maketitle              
\begin{abstract}
Hydropower plants are crucial for stable renewable energy and serve as vital water sources for sustainable agriculture. However, it is essential to assess the current water management practices associated with hydropower plant management software. A key concern is the potential conflict between electricity generation and agricultural water needs. Prioritising water for electricity generation can reduce irrigation availability in agriculture during crucial periods like droughts, impacting crop yields and regional food security. Coordination between electricity and agricultural water allocation is necessary to ensure optimal and environmentally sound practices.
Neural networks have become valuable tools for hydropower plant management, but their black-box nature raises concerns about transparency in decision making. Additionally, current approaches often do not take advantage of their potential to create a system that effectively balances water allocation.

This work is a call for attention and highlights the potential risks of deploying neural network-based hydropower plant management software without proper scrutiny and control. To address these concerns, we propose the adoption of the Agriculture Conscious Hydropower Plant Management framework, aiming to maximise electricity production while prioritising stable irrigation for agriculture. We also advocate reevaluating government-imposed minimum water guidelines for irrigation to ensure flexibility and effective water allocation. Additionally, we suggest a set of regulatory measures to promote model transparency and robustness, certifying software that makes conscious and intelligent water allocation decisions, ultimately safeguarding agriculture from undue strain during droughts.

\end{abstract}
\section{Introduction}

Hydropower plants play a vital role in our renewable energy infrastructure by harnessing the kinetic energy of flowing water to generate electricity. The efficiency of these power plants is often enhanced through the construction of reservoirs and dams, which serve as essential components for controlling and optimising the energy derived from water sources. Dams serve as effective barriers that impound substantial volumes of water within reservoirs, creating a regulated water supply.

In addition to electricity generation, some reservoirs also serve as vital water sources that support irrigation systems to promote sustainable agriculture.
The provision of water for irrigation is of paramount significance, given its pivotal role in ensuring global food security and supporting agriculture, which forms the backbone of numerous economies. Irrigation systems are instrumental in increasing natural rainfall and meeting the water requirements of crops, particularly in regions characterised by erratic or insufficient precipitation. Effective irrigation practices not only enhance crop yields, but also facilitate the cultivation of a wider variety of crops and allow year-round agricultural production. Consequently, these measures contribute to stabilising food production, increasing food availability, and reducing the susceptibility of communities to food shortages and price fluctuations \cite{tilmantAgriculturaltohydropowerWaterTransfers2009}.

Governments play a crucial role in regulating the availability of irrigation water by setting minimum water guidelines for reservoirs associated with hydropower plants. These guidelines are formulated to address a range of critical objectives, including the preservation of agricultural productivity, which is fundamental for ensuring food security and economic stability. Moreover, these regulations are designed to safeguard the natural ecosystems and aquatic life within the reservoir and downstream areas. Failure to meet these minimum guidelines leads to fines applied to the respective management company \cite{schafferModellingEnvironmentalConstraints2020}.

Traditionally, hydropower plant management has relied heavily on software solutions that incorporate traditional optimisation algorithms. These software tools are designed to ensure the efficient utilisation of available water resources for electricity generation, thus making real-time decisions about water release, reservoir levels, and electricity generation. These tools analyse various factors such as water inflow forecasts, electricity demand, and environmental constraints to determine the most advantageous operational strategies \cite{schafferModellingEnvironmentalConstraints2020,fengMultiobjectiveOperationOptimization2017}. 

With the success of Neural Networks (NNs) as powerful predictors, several works in the literature have shown NNs can be powerful tools for hydropower plant management \cite{abdulkadir2012application,linRevenuePredictionIntegrated2022}. However, concerns arise due to their inherent black-box nature, lacking transparency in decision making. This raises doubts about the adherence to minimum water guidelines for maintaining agricultural and ecosystem preservation water levels, posing threats to food security and natural life.
To address potential fines and regulatory issues, one strategy is to deduct minimum water guidelines required for agriculture and ecosystem preservation from the inflows before giving them as input to the NN, prioritising energy production, and overlooking optimising agricultural water usage. 

In both traditional optimisation-based and NN-based software, the predominant training objective frequently centres around the maximisation of revenue, possibly overlooking the broader considerations of sustainability and social aspects in water resource management. This is achieved through the definition of objective functions that primarily reward electricity generation or revenue without adequately promoting prudent and strategic water storage for irrigation purposes. 
Given the critical role of hydropower plants in ensuring stable renewable energy and their importance as vital water sources for sustainable agriculture, it is essential to assess the current water management practices associated with these systems. A key concern is the potential conflict between electricity generation and agricultural water needs. Prioritising water for electricity generation can reduce the availability of irrigation water during crucial periods such as droughts, impacting crop yields and regional food security. Coordination between energy and agricultural water allocation is necessary to ensure optimal and environmentally sound practices.

In this work, our main goal is to draw attention to the potential harms associated with deploying NN hydropower plant management software in production without proper scrutiny and control. We also aim to highlight the missed opportunity to integrate electricity generation with agriculture sustainability and food security. To address these concerns, we propose the adoption of the Agriculture Conscious Hydropower Plant Management framework (ACHPM) as a novel approach for developing these software. 
 The importance of our research lies in recognising that the uncontrolled and unregulated deployment of NN-based hydropower management software may have adverse consequences on both economic and environmental fronts. Furthermore, current practices fail to take advantage of the full capabilities of NNs. By highlighting the importance of this issue, we hope to inspire a shift towards more responsible and sustainable practices in the field of hydropower management with NN models.

The structure of this paper is outlined as follows. In Section \ref{sec:background}, we provide a brief exploration of approaches to hydropower plant management, using traditional optimisation techniques, and more contemporary approaches, using NNs.
Section \ref{sec:method} delves into the proposed ACHPM framework and underscores the importance of regulating and scrutinising black-box models before their deployment.
Finally, in Section \ref{sec:conclusion} we conclude this paper by summarising the key takeaways and insights and discuss some future work.

\section{Hydropower Plant Management} \label{sec:background}

Hydropower plants are crucial for renewable energy generation and some play an equally important role as a water source for agriculture. Thus, the efficient management of hydropower plants is crucial to optimise their performance, ensure a stable electricity supply to consumers, preserve the surrounding ecosystem, and promote agricultural sustainability.
Over time, companies responsible for hydropower plants have relied on software solutions for management. Initially, these software solutions were based on traditional constrained optimisation techniques aimed at maximising total revenue while satisfying specific constraints \cite{grygierAlgorithmsOptimizingHydropower1985}. Note that many software still use this type of approach \cite{zhangChanceconstrainedCooptimizationDayahead2022}.

However, with the rise of NNs and their remarkable performance across various tasks, the landscape of software for hydropower plant management is undergoing a shift towards NN-based solutions, as evidenced by various examples in the literature \cite{barzola-montesesHydropowerProductionPrediction2022,bordinMachineLearningHydropower2020}. In this section, we provide a brief overview and a general mathematical formulation of the management problem when solved using constrained optimisation techniques and NNs. We also emphasise the distinctions between these two approaches and delineate the challenges associated with each strategy.

\subsection{Constrained Optimisation Approach}

The problem of managing a hydropower plant is generally formulated as the constrained optimisation problem \eqref{eq:p1} in which the goal is to maximise an objective function $l(\boldsymbol{\theta})$, typically defined by the amount of electricity generated, \emph{i.e.} the company's profit, while satisfying some constraints \cite{fengMultiobjectiveOperationOptimization2017,grygierAlgorithmsOptimizingHydropower1985}. 
In this work, we focus on the constraints related to meeting the electricity demand and minimum water guidelines for irrigation and ecosystem preservation. In general, the formulation of the optimisation problem is defined by:

\begin{maxi}|l|[2]
    {\boldsymbol{\theta} \in \mathbb{R}^{n_{\boldsymbol{\theta}}}}{l(\boldsymbol{\theta}),}
    {\label{eq:p1}}
    {}
    \addConstraint{P(\boldsymbol{\theta})}{\geq P_{demand}}
    \addConstraint{Q_{river}(\boldsymbol{\theta})}{\geq Q^{\min}_{river}} 
    \addConstraint{Q_{irrigation}(\boldsymbol{\theta})}{\geq Q^{\min}_{irrigation}},
\end{maxi}

\noindent where $\boldsymbol{\theta} \in \mathbb{R}^{n_{\boldsymbol{\theta}}}$ are the parameters to optimise, $l$ is the objective function, $P$ is the electricity generated, $P_{demand}$ the electricity required to meet consumer needs, $Q_{river}$ the amount of water for the preservation of ecosystems, $Q^{\min}_{river}$ the minimum water guideline for ecosystem preservation, $Q_{irrigation}$ the amount of water allocated for irrigation and $Q^{\min}_{irrigation}$ the minimum water guideline for irrigation.

In general, a common behaviour observed when solving \eqref{eq:p1} is that traditional optimisation algorithms' solution only allocate the minimum water guidelines for irrigation and ecosystem preservation since the objective is to maximise electricity generation. Since there is no direct incentive or reward for allocating more water to these other purposes, in the mathematical formulation \eqref{eq:p1}, the optimal solution prioritise electricity production. Allocating strictly the minimum water guidelines for irrigation and ecosystem preservation allows for more water to be available for electricity generation, which, in turn, increases the value of the objective function.

This outcome underscores a potential limitation of such optimisation approaches, as they may not inherently account for suitable broader social and environmental considerations associated with water resource management.  

\subsection{Neural Networks Approach}

Analysing a NN training process from the perspective of traditional optimisation, one can draw parallels to an algorithm for solving the unconstrained problem \eqref{eq:p2}. This view is based on the fact that during NN training, the primary objective is to minimise an objective function $l(\boldsymbol{\theta})$, that measures the error between the predicted and ground-truth values. 

The optimisation problem is given by:
 \begin{mini}|l|[2]
    {\boldsymbol{\theta} \in \mathbb{R}^{n_{\boldsymbol{\theta}}}}{l(\boldsymbol{\theta}),}
    {\label{eq:p2}}
    {}
\end{mini}

\noindent where $l$ is the objective function used in NN training, commonly referred to as the loss function. $\boldsymbol{\theta} \in \mathbb{R}_{n_{\boldsymbol{\theta}}}$ are the NN parameters to be optimised. 

In general, in this type of approach, there are not constraints on the parameters of the problem \eqref{eq:p2}. Furthermore, if necessary, incorporating constraints can be nontrivial and may not align with the standard practice of NN training.

The NN approach has been extensively used for hydropower plant management in which the goal is mainly to predict electricity generation given a set of input data \cite{barzola-montesesHydropowerProductionPrediction2022,bordinMachineLearningHydropower2020}. However, few works can also be found in the literature that solve the problem incorporating constraints on parameters $\boldsymbol{\theta}$ \eqref{eq:p1} \cite{shawHydropowerOptimizationUsing2017}. 

To the best of our knowledge, there has been a significant gap between optimising irrigation water allocation and electricity generation. The existing approaches, which focuses primarily on electricity generation and profit maximisation, are limited in their capacity to make conscious, balanced decisions regarding water management for irrigation. This limitation poses risks to society as it does not account for the critical role that water plays in supporting agriculture and food security.

Expanding the NN framework to include the optimisation of irrigation water in the loss function is a crucial step towards addressing these limitations. This approach would enable more informed and responsible decision making in water resource management, allowing for a comprehensive assessment of trade-offs between electricity generation and the needs of agriculture, thus promoting a more sustainable and equitable allocation of this valuable resource.

\section{The Proposal} \label{sec:method}

In this section, we introduce the ACHPM framework as an innovative approach aimed at conscious and balanced hydropower plant management. The ACHPM framework seeks to address the limitations of existing software by integrating ethical considerations, promoting sustainability, and providing a more comprehensive model for managing water.

Additionally, we emphasise the importance of regulating and scrutinising these software solutions before they enter the production stage. Such regulatory measures are essential to ensure that the software is not only effective but also ethical and transparent. By taking these steps, we can better tackle the challenges posed by climate change, which directly impact agriculture and food security. This approach reflects a commitment to responsible and sustainable management of our valuable water resources and their crucial role in addressing the pressing challenges of our time.

\subsection{The Agriculture Conscious Hydropower Plant Management Framework}

As discussed previously, existing hydropower management approaches predominantly prioritise maximising electricity production, offering no explicit incentives to allocate more water for irrigation, indirectly penalising such allocations. Given the current era marked by climate change, which alters the seasonality of ecosystems and introduces instability in rainfall patterns essential for crop growth, it has become increasingly clear that this management approach is inadequate and overly simplistic.

The existing state of management software in many hydropower plant management companies typically adheres to government-imposed minimum water guidelines. This rigid approach underscores the missed opportunity to intelligently allocate water for irrigation based on predictive models that account for the likelihood of dry or wet periods. The proposed ACHPM framework seeks to address this deficiency by advocating for an approach that not only maximises electricity production but also prioritises the provision of a stable source of irrigation for agriculture. By anticipating higher or lower irrigation needs and adjusting water allocation accordingly, this framework aims to establish a resilient water supply for agriculture, especially during drought periods that can lead to crop loss, food scarcity, rising food prices, and, ultimately, food poverty.

\begin{figure}
    \centering
    \includegraphics[width=\textwidth]{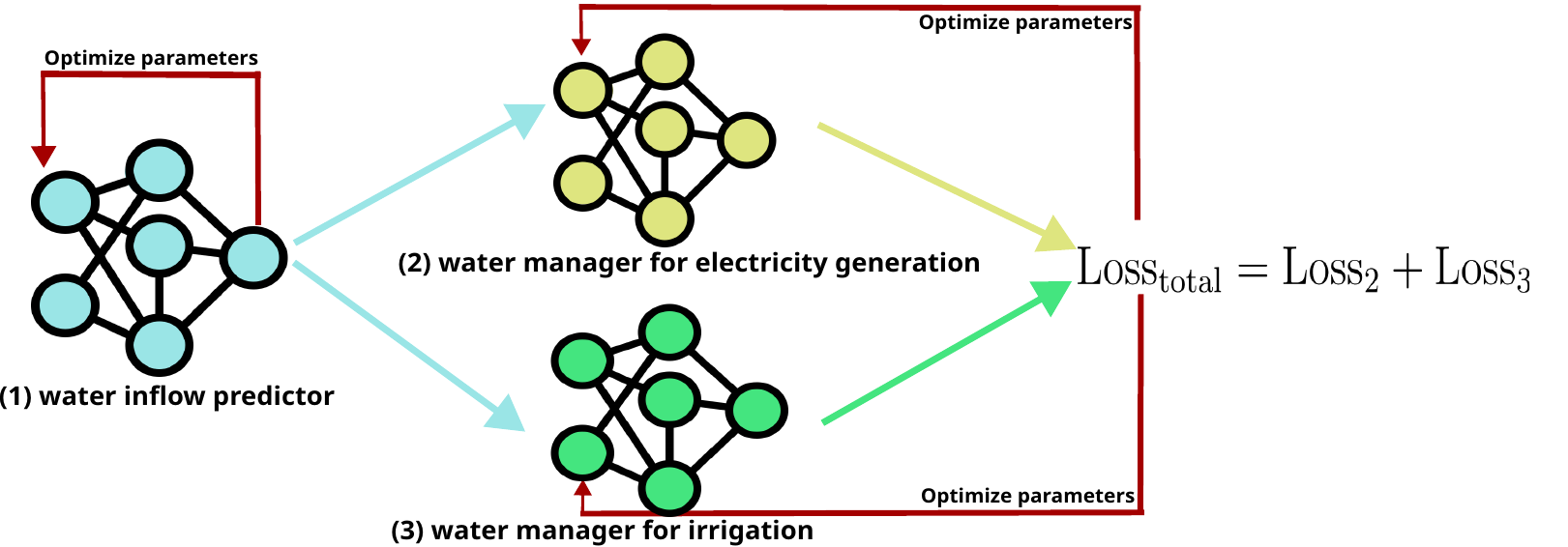}
    \caption{The ACHPM framework training process.}
    \label{fig:ACHPMframework}
\end{figure}

The ACHPM framework, Figure \ref{fig:ACHPMframework}, comprises an ensemble of NNs organised into three components with different goals:

\begin{enumerate}[label={(\arabic*)}]
    \item \textbf{Water inflow predictor:} A NN to forecast reservoir water inflow using historical data and climate patterns, enabling anticipation of water availability during dry and wet seasons. The information given by this NN forms the foundation for informed decision making in hydropower management;

    \item \textbf{Water manager for electricity generation:} A NN tasked with optimising the allocation of water for electricity generation. It takes into account the predictions of water inflows provided by the water inflow predictor NN, along with other input data such as electricity demand, turbine efficiency, operational constraints, among others;

    \item \textbf{Water manager for irrigation:} A NN dedicated to optimising the allocation of water for crop irrigation. It relies on the forecasts of water inflows made by the water inflow predictor NN, in addition to input data such as soil moisture levels, atmospheric conditions, and the water needs of the targeted crops. The objective here is to allocate water resources efficiently to support agricultural sustainability.
\end{enumerate}

The training approach for the ensemble of neural networks in the ACHPM framework involves a two-stage process:
\begin{itemize}
    \item \textit{First Stage - water inflow predictor training:} In this initial stage, the focus is on optimising the water inflow predictor NN. This NN is trained using the loss function \eqref{eq:p2}, that measures the error between the predicted and the ground-truth values, enabling the NN to learn a model that accurately captures the patterns of water inflows over time;

    \item \textit{Second Stage - water manager for electricity generation and irrigation training:}  The second stage is characterised by the joint training of the water manager for electricity generation and the water manager for irrigation NNs. Both of these NNs share a loss function $\text{Loss}_\text{total}$. Both NNs receive as input, the output of the water inflow predictor NN and specific input data for each NN. The two NNs must balance water resource allocation to minimise the shared loss function. It important to note that, in this stage, the reward is not solely based on electricity generation. Instead, it is extended to include the goal of securing sufficient irrigation water to meet demand over time. The shared loss function guides the training process by having the contributions of each NN ensuring maximisation of electricity generation, $\text{Loss}_2(\boldsymbol{\theta}_2)$, and clever water allocation for irrigation, $\text{Loss}_3(\boldsymbol{\theta}_3)$. $\boldsymbol{\theta}_2,\boldsymbol{\theta}_3$ are the parameters of the water manager for electricity generation and for irrigation NNs, respectively.
    
%
    
\end{itemize}

This two-stage training approach equips the ACHPM framework with the capacity to make informed and balanced decisions in hydropower management, considering both economic and agricultural objectives while predicting and adapting to changing environment conditions.

It should be noted that, upon initial examination, it might appear that implementing such a framework would result in reduced water availability for electricity generation. However, the existing practice of maintaining minimum water guidelines allocations, even during wet periods, is often unnecessary and could be used more efficiently to benefit both agriculture and the profitability of hydropower companies.

One of the concerns associated with using NNs in hydropower management is the perception that the traditional training process resembles an unconstrained optimisation problem. Furthermore, the black-box nature of NNs can make it challenging to ensure that critical constraints, such as minimum water guidelines for ecosystem preservation, are consistently maintained. This is a significant issue as violating these constraints may pose a risk to natural life and lead to fines imposed by governments on management companies.

To address this challenge, we propose an approach in which the constraints are explicitly incorporated into the ACHPM framework. This can be achieved using established methods found in the literature that enable to solve a constrained optimisation problem during NN training. By explicitly integrating these constraints into the training process, we can ensure that the NNs are not only optimised for their primary objectives but also guarantee that the constraints are consistently satisfied \cite{coelho2023prior}.

\subsection{Software Regulation}

In addition to presenting the ACHPM framework, we strongly advocate for the implementation of regulations governing hydropower plant management software. The development of this framework has been influenced by the shortcomings of existing approaches in achieving a balance between electricity generation and the allocation of water for irrigation. Our advocacy centres on the need to reconsider the current reliance on minimum water guidelines for irrigation imposed by governments, as they often introduce inflexibility and may be exploited by companies, leading to suboptimal water allocation for agriculture.

We propose the adoption of regulations that mandate management software to prioritise conscious and adaptive water allocation for irrigation. This approach requires management software to respond to the actual needs of agriculture by successfully adjusting the allocation of water as conditions demand. In doing so, these software solutions can actively promote agricultural sustainability and contribute to a stable food supply, ensuring the well-being of populations. This shift in focus reflects a commitment to a more responsive and equitable approach to hydropower management in the face of changing environmental conditions and evolving societal needs.

To ensure the successful implementation of hydropower management software that prioritises conscious water allocation for irrigation and electricity generation, transparency and accountability are vital. We advocate regulating entities to promote:

\begin{itemize}
    \item \textbf{Transparency:} It is imperative that the architectures and training procedures of the NNs used in these software solutions are transparent and well documented. This transparency allows for the validation of imposed requirements;

    \item \textbf{Evaluation:} The creation of robust evaluation metrics is essential for assessing the performance of the software in real-world scenarios. These metrics should measure the software's ability to predict drought periods and efficiently allocate water, penalising instances where the software pressures the irrigation system when it could have allocated more water for irrigation during unconstrained electricity generation periods;

    \item \textbf{Regular auditing and compliance checks:} Due to the high variability of atmospheric conditions due to climate change, to ensure that the management software continues to adhere to the requirements, regular auditing and compliance checks should be conducted. This prevents outdated software and helps holding management companies accountable for their software's performance.
\end{itemize}

These measures collectively contribute to the responsible and sustainable management of water resources, achieving the dual goals of efficient electricity production and robust support for agriculture, especially during critical periods such as droughts. This approach reflects a commitment to addressing the challenges of climate change while ensuring the availability of essential resources like a stable food supply.

\section{Conclusion and Future Work} \label{sec:conclusion}

This work is a call for attention to the impact of current and future NN-based approaches for hydropower plant management software in agriculture. We highlighted the common goal of management companies to maximise the electricity generation while limiting the allocation of water to irrigation to the minimum guidelines issued by the government as to avoid fines. 

The impact of climate change on water availability and agricultural sustainability requires a more sophisticated and adaptive approach to hydropower management. Such an approach should be based on broader social needs, including agriculture and food security. It is crucial to move beyond the naive optimisation of electricity production and embrace a more holistic and environmentally conscious strategy that recognises the intricate interplay between hydropower and agriculture in the face of climate variability.

To fix this issue, we proposed the ACHPM framework, designed to promote responsible software capable to balance water resource allocation. Thus, the proposed framework is an ensemble of three NNs working together. The first NN acts as a water inflow predictor, modelling the behaviour of the inflows over time giving the ability of predicting dry and wet periods. The second and third NNs are trained jointly using a shared loss function in which one dedicated to managing water for electricity generation and the other for irrigation. The shared loss function must be chosen so that the models balance the allocation of water resources. 
Given the inherent black-box nature of NNs and their limitations in guaranteeing constraint satisfaction, such as the minimum water guideline for ecosystem preservation, we suggest using an approach that explicitly introduces constraints during NN training.

Furthermore, this work underscores the need for regulations governing hydropower management software, including evaluation and certification processes by regulatory entities prior to production. It emphasises the importance of transparency in model architectures and training procedures to ensure that software objectives align with societal needs rather than just profit maximisation. Creating evaluation metrics to assess the software's ability to balance water allocation and conducting regular auditing and compliance checks are also crucial steps in promoting responsible management.

As future work, we encourage the practical application of the ACHPM framework to real data and performance analysis to refine and enhance the framework's effectiveness. In addition, the development of shared loss functions that further promote balanced water usage and the establishment of comprehensive evaluation metrics remain open challenges to ensure the robustness of such systems in addressing the complex interplay of electricity generation and agriculture sustainability.

\section*{Acknowledgments}
The authors acknowledge the funding by Fundação para a Ciência e Tecnologia (Portuguese Foundation for Science and Technology) through CMAT projects UIDB/00013/2020 and UIDP/00013/2020 and the funding by FCT and Google Cloud partnership through projects CPCA-IAC/AV/589164/2023 and CPCA-IAC/AF/589140/2023.

\noindent C. Coelho would like to thank FCT the funding through the scholarship with reference 2021.05201.BD.
L.L. Ferrás would also like to thank the funding by FCT through the project 2022.06672.PTDC

\bibliographystyle{splncs04}
\bibliography{mybibliography}

\end{document}